\documentclass[twoside,11pt]{article}

\usepackage{jmlr2e}
\usepackage{booktabs}
\usepackage{paralist}
\usepackage{graphicx}
\usepackage{adjustbox} 

\usepackage{lastpage}

\usepackage{listings}
\usepackage{minted}

\ShortHeadings{}{Gomes, Lee, Gunasekara, Sun, Cassales, Liu, Heyden, Cerqueira, et al.}
\firstpageno{1}

\begin{document}

\title{CapyMOA: \\Efficient Machine Learning for Data Streams and Online Continual Learning in Python}

\author{\name Heitor Murilo Gomes \email heitor.gomes@vuw.ac.nz \\
       \addr Victoria University of Wellington, New Zealand
       \AND
       \name Anton Lee \email anton.lee@vuw.ac.nz \\
       \addr Victoria University of Wellington, New Zealand
       \AND
       \name Nuwan Gunasekara \email nuwan.gunasekara@hh.se \\
       \addr Hamlstad University, Sweden
       \AND
       \name Yibin Sun \email yibin.spencer.sun@gmail.com \\
       \addr AI Institute, University of Waikato, New Zealand
       \AND
       \name Guilherme Weigert Cassales \email guilherme.cassales@waikato.ac.nz \\
       \addr AI Institute, University of Waikato, New Zealand
       \AND
       \name Jia Justin Liu \email justin.l@waikato.ac.nz \\
       \addr AI Institute, University of Waikato, New Zealand
       \AND
       \name Marco Heyden \email marco.heyden@kit.edu \\
       \addr Karlsruhe Institute of Technology, Germany
       \AND
       \name Vitor Cerqueira \email vcerqueira@fe.up.pt \\
       \addr University of Porto, Portugal
       \AND
       \name Maroua Bahri \email maroua.bahri@lip6.fr \\
       \addr Sorbonne Université, CNRS, LIP6, France
       \AND
       \name Yun Sing Koh \email y.koh@auckland.ac.nz \\
       \addr University of Auckland, New Zealand
       \AND
       \name Bernhard Pfahringer \email bernhard@waikato.ac.nz \\
       \addr AI Institute, University of Waikato, New Zealand
       \AND
       \name Albert Bifet \email abifet@waikato.ac.nz \\
       \addr AI Institute, University of Waikato, New Zealand
}
\editor{}

\maketitle

\begin{abstract}
CapyMOA is an open-source Python library for efficient machine learning on data streams and online continual learning.
It provides a structured framework for real-time learning, supporting adaptive models that evolve over time.
CapyMOA's architecture allows integration with frameworks such as MOA, scikit-learn and PyTorch, enabling the combination of high-performance online algorithms with modern deep learning techniques.
By emphasizing efficiency, scalability, and usability, CapyMOA allows researchers and practitioners to tackle dynamic learning challenges across various domains.
Website: \url{https://capymoa.org}. GitHub: \url{https://github.com/adaptive-machine-learning/CapyMOA}. 
\end{abstract}


\section{Introduction}
CapyMOA is an open-source Python library (BSD 3-Clause License) for efficient stream learning and online continual learning (OCL), actively used by a growing community of researchers and practitioners.
It provides a structured and extensible framework that enables real-time model updates, while ensuring \textbf{efficiency}, \textbf{interoperability}, and \textbf{accessibility}.
By offering a diverse set of algorithms and tools, CapyMOA supports both research and practical applications. 
CapyMOA supports a wide range of online use cases, including classification, regression, anomaly detection, concept drift detection, clustering, semi-supervised learning and OCL~\citep{aljundi2019gradient}.
CapyMOA goes beyond existing libraries by offering a modern high-level Python API coupled with efficient implementations. 

\section{Background and Related Work}
Machine learning for data streams presents unique challenges that distinguish it from traditional batch learning~\citep{bifet2018machine,gomes2019machine}. Unlike static datasets, streaming data arrives continuously and must be processed incrementally without storing the entire dataset. 
Two key challenges in such scenarios are \textbf{concept drift}~\citep{gama2014survey}, where the statistical properties of the data change over time, requiring models to adjust dynamically to maintain predictive performance, and \textbf{computational constraints}, as models must process each instance upon arrival while operating under strict memory and processing limitations. To effectively address these challenges, practitioners and researchers need specialized tools for evaluating, analyzing, and experimenting with various algorithms.

Several frameworks (e.g., MOA~\citep{ref_moa}, Scikit-Multiflow~\citep{ref_skmf}, and River~\citep{ref_river}) have been developed to address stream learning.

\textbf{MOA} is a well-established Java-based framework with a large range of efficient implementations, but its Java-based code limits accessibility for new users that often prefer Python. \textbf{River} is a fully Python-based framework that focuses on usability and flexibility but sacrifices computational efficiency due to its pure Python implementation. \textbf{Scikit-Multiflow} was initially developed to provide Python implementations of MOA algorithms but has since been largely replaced by River. MOA offers efficient stream learning algorithms, but it lacks the accessibility of a Python-native tool. Conversely, River prioritizes usability but struggles with performance, highlighting a trade-off between efficiency and accessibility in existing frameworks. In parallel, frameworks such as \textbf{Avalanche}~\citep{carta2023avalanche} have been developed for continual learning, providing support for OCL scenarios based on streams of experiences. However, Avalanche operate at the level of dataset sequences rather than instance-level streams, and do not directly address challenges such as concept drift or strict computational constraints, hence it misses opportunities of integrating data stream learning with OCL~\citep{gunasekara2023survey}. 

CapyMOA addresses these limitations by providing an efficient, extensible, and user-friendly framework for stream learning and OCL. It is built upon three core pillars: 
\textbf{Efficiency}, leveraging optimized implementations with minimal memory overhead in a modern Python framework; \textbf{Interoperability}, supporting both native Python implementations and seamless integration with MOA and PyTorch for hybrid learning; and \textbf{Accessibility}, providing an intuitive Python API that simplifies complex stream learning and OCL tasks, such as concept drift detection and class-incremental problems~\citep{de2021continual}. 

Furthermore, none of the existing frameworks provide a unified API that bridges stream learning and OCL, allowing both paradigms to naturally intersect and benefit from each other.



\section{Core Features of CapyMOA}

\subsection{Data Representation and Interoperability}
CapyMOA provides a structured approach to handling data streams through its \textbf{Schema} and \textbf{Instance} representations. A \textbf{Schema} defines the structure of a data stream, specifying the number of attributes, their types (nominal or numeric), and the target variable. 
An \textbf{Instance} represents a single data point within a stream, adhering to a Schema. Unlike other frameworks~\citep{ref_river} that use flexible but potentially ambiguous dictionary-based representations, CapyMOA Schemas eliminate silent errors caused by inconsistent feature representations. This design enhances stability, making CapyMOA particularly well-suited for long-term stream learning tasks. Importantly, the Instance abstraction is polymorphic, allowing the same interface to support different underlying data representations. In particular, Instances can be seamlessly consumed by both traditional learners and tensor-based models (e.g., PyTorch), enabling smooth interoperability without requiring users to manually convert data within the training loop. 
Furthermore, CapyMOA integrates with high-performance Java-based backends (e.g., MOA) via a lightweight bridging layer (JPype), allowing efficient cross-language interaction while preserving a consistent Python-level abstraction.

\subsection{Concept Drift: Simulation, Detection and Evaluation}
CapyMOA introduces a unified framework for \textbf{simulating} concept drift through a simple declarative specification of stream segments and drift events. For example, a stream can be defined as a sequence such as 
\[[SEA(function=1), AbruptDrift(position=5000), SEA(function=2)]\] which can be extended to model recurrent drift patterns. 
CapyMOA includes both classic univariate \textbf{detectors} such as ADWIN~\citep{adwin} and modern multivariate approaches such as ABCD~\citep{abcd}, enabling flexible drift detection across different settings. 
Finally, CapyMOA supports an \textbf{evaluation} interface for assessing drift detection performance based on detection delay and detection-based metrics, allowing systematic comparison of strategies. 

\subsection{Online Continual Learning (OCL)}
CapyMOA offers native support for \textbf{Online Continual Learning} by integrating task-based learning within the same abstractions used for data streams. Rather than introducing a separate workflow, CapyMOA enables OCL through shared representations and training interfaces, allowing both instance-based and mini-batch learners to operate seamlessly across streaming and task-based settings. 

CapyMOA supports standard continual learning scenarios, including task-incremental and class-incremental settings, with explicit control over task boundaries and data streams. It provides a unified evaluation interface through the \texttt{ocl\_train\_eval\_loop}, which extends prequential evaluation to account for performance across tasks and over time, including anytime evaluation. This design allows stream learning methods and continual learning strategies to be combined within the same system, enabling hybrid approaches that leverage drift detection, replay mechanisms, and adaptive models in a consistent framework.

\subsection{Efficiency and Scalability}
CapyMOA combines high-performance backend implementations with the flexibility of a modern Python interface. To evaluate efficiency, we compare CapyMOA with MOA, River, and Scikit-Multiflow using Streaming Random Patches (SRP)~\citep{ref_srp} (30 base learners), a widely used and computationally representative benchmark in stream learning, on a synthetic Hyperplane dataset with $10^5$ instances\footnote{additional benchmarks are available at \url{https://github.com/adaptive-machine-learning/CapyMOA}.}.


\begin{table}[!h]
\centering
\caption{\textsc{Runtime of SRP$_{30}$ on Hyper100k Dataset}}
\label{tb:srp30_time}
\begin{adjustbox}{max width=\linewidth}
\begin{tabular}{lc} 
    \toprule
    \textsc{Platforms} & \textsc{Runtime (s)} \\
    \midrule
    MOA & 35.81 ± 0.70 \\ 
    CapyMOA & 34.14 ± 0.86 \\ 
    Scikit-multiflow & 4,524 ± 1,387 \\ 
    River & 1,396 ± 51,08 \\ 
    \bottomrule
\end{tabular}
\end{adjustbox}
\end{table}

As shown in Table~\ref{tb:srp30_time}, CapyMOA achieves performance comparable to MOA while significantly outperforming Python-based alternatives in runtime, demonstrating that efficiency can be preserved without sacrificing usability.

\section{Conclusion}
CapyMOA provides a unified and efficient framework for machine learning on data streams and online continual learning, combining strong abstractions with high-performance implementations. CapyMOA supports a wide range of tasks including classification, anomaly detection, clustering and semi-supervised learning through a consistent API, with detailed tutorials and documentation available at \url{https://capymoa.org}, including how to contribute new algorithms (\url{https://capymoa.org/contributing/}). The library is supported by a comprehensive testing framework, including extensive unit tests and continuous integration, ensuring reliability and reproducibility as the system evolves.

Future work includes expanding support for OCL methods, improving drift evaluation, and enhancing deep learning integration. We expect CapyMOA to continue to evolve as a comprehensive tool for both research and real-world applications.


\bibliography{paper}

\end{document}